\documentclass[dvipsnames,letterpaper, 10 pt, conference]{ieeeconf}

\IEEEoverridecommandlockouts	

\usepackage{cite}
\usepackage{amsmath,amssymb,amsfonts}
\usepackage{algorithmic}
\usepackage{textcomp}
\usepackage{graphicx,color}
\usepackage{mathrsfs}
\usepackage{tikz}
\usepackage[vlined,ruled]{algorithm2e}
\usepackage{subfigure}
\usepackage{url}
\usepackage{color}
\usepackage{dsfont}
\usepackage{bbm}
\usepackage{booktabs}
\usepackage{array}
\usepackage{xcolor}
\usepackage{yfonts}
\usepackage[normalem]{ulem}
\usepackage{arydshln,leftidx,mathtools}
\usepackage{multicol}
\usepackage{amsmath}
\usepackage{stfloats}
\usepackage{comment}
\usepackage{subfigure}
\usepackage{manyfoot}
\usepackage{listings}
\usepackage{float}
\usepackage[most]{tcolorbox}
\usepackage{placeins}
\usetikzlibrary{positioning,arrows.meta,shapes.misc}
\usepackage{pifont} 

\definecolor{codegreen}{rgb}{0,0.6,0}
\definecolor{codegray}{rgb}{0.5,0.5,0.5}
\definecolor{codepurple}{rgb}{0.58,0,0.82}
\definecolor{backcolour}{rgb}{0.95,0.95,0.92}

\lstdefinestyle{mystyle}{
    backgroundcolor=\color{backcolour},   
    commentstyle=\color{codegreen},
    keywordstyle=\color{magenta},
    numberstyle=\tiny\color{codegray},
    stringstyle=\color{codepurple},
    basicstyle=\ttfamily\footnotesize,
    breakatwhitespace=false,         
    breaklines=true,                 
    captionpos=b,                    
    keepspaces=true,                 
    numbers=left,                    
    numbersep=5pt,                  
    showspaces=false,                
    showstringspaces=false,
    showtabs=false,                  
    tabsize=2
}

\definecolor{myBlue}{RGB}{49,130,189}

\DontPrintSemicolon
\SetKw{KwContinue}{continue}
\SetKw{KwAnd}{and}
\SetKw{KwRet}{return}
\SetKwInput{KwProc}{Procedure}

\usepackage{tabularx}
\usepackage{multirow}
\usepackage{makecell}

\newcommand\aamsout{\bgroup\markoverwith{\textcolor{violet}{\rule[0.5ex]{2pt}{1pt}}}\ULon}

\DeclareSymbolFont{bbold}{U}{bbold}{m}{n}
\DeclareSymbolFontAlphabet{\mathbbold}{bbold}

\newcommand\oprocendsymbol{\hbox{$\square$}}
\newcommand\oprocend{\relax\ifmmode\else\unskip\hfill\fi\oprocendsymbol}

\tcbset{
  panel/.style n args={2}{ 
    enhanced,
    arc=10pt,                 
    boxrule=0.5pt,            
    colframe=#1!80!black,
    colback=#1!12,            
    colbacktitle=#1!25,       
    coltitle=black,
    fonttitle=\bfseries\normalsize,
,
    title={#2},
    title filled,             
    titlerule=1.2pt,          
    toptitle=2mm, bottomtitle=2mm,
    left=10pt,right=10pt,top=10pt,bottom=10pt,
    before skip=8pt, after skip=10pt,
  }
}

\newtcolorbox{TaskBox}[1]{panel={blue}{#1}}
\newtcolorbox{EnvBox}[1]{panel={green}{#1}}

\newtcolorbox{StageBox}{
  enhanced,
  colback=white,
  colframe=black!60,
  boxrule=0.8pt,
  arc=6pt,
  left=8pt,right=8pt,top=6pt,bottom=6pt,
  before skip=6pt, after skip=8pt,
}

\newcommand*{\QEDA}{\hfill\ensuremath{\blacksquare}}%

\DeclareNewFootnote{R}[roman]

\graphicspath{{figs/}}

\makeatletter
\let\NAT@parse\undefined
\makeatother
\usepackage[colorlinks,urlcolor=blue, linkcolor=blue, citecolor=blue]{hyperref}

\begin{document}

\title{\LARGE \bf Test-Driven Agentic Framework for Reliable Robot Controller}

\author{Shivanshu~Tripathi, Reza~Akbarian~Bafghi, and
  Maziar~Raissi \thanks{S.~Tripathi is with the Department of Electrical and Computer Engineering, and M.~Raissi is with the Department of Mathematics at the University of California, Riverside, \href{mailto:strip008@ucr.edu}{\{\texttt{strip008}},\href{mailto:maziarr@ucr.edu}{\texttt{maziarr\}@ucr.edu}}. R.~A.~Bafghi is with the Department of
    Computer Science, University of Colorado, Boulder, \href{mailto:reak3132@colorado.edu}{\{\texttt{reak3132@colorado.edu}}\}. }}

\maketitle
\pagestyle{empty}
\thispagestyle{empty}

\begin{abstract}
In this work, we present a test-driven, agentic framework for synthesizing a 
deployable low-level robot controller for navigation tasks. Given a 2D map with an 
image of an ultrasonic sensor-based robot, or a 3D robotic simulation environment, 
our framework iteratively refines the generated controller code using diagnostic 
feedback from structured test suites to achieve task success.
We propose a dual-tier repair strategy to refine the generated code that alternates between 
prompt-level refinement and direct code editing. 
We evaluate the approach across 2D navigation tasks and 3D 
navigation in the Webots simulator. Experimental results show that 
test-driven synthesis substantially improves controller reliability and 
robustness over one-shot controller generation, especially when the initial 
prompt is underspecified. The source code and 
demonstration videos are available at: 
\url{https://shivanshutripath.github.io/robotic_controller.github.io/}.
  \end{abstract}


\section{Introduction}\label{sec: introduction}
Motion planning is fundamental in robotics, requiring autonomous 
systems to generate collision-free, and dynamically feasible trajectories 
\cite{SML:06,SML-JJK:01}. However, autonomy 
hinges on execution, i.e., a controller must follow the plan while 
adhering to the robot's sensing, actuation, and environmental constraints. 
This execution layer is frequently the primary source of brittleness. Sensors 
have discretized resolution and limited fields of view, 
and actuators are subject to saturation and rate limits. Consequently, a 
significant gap exists between the theoretical existence of a plan 
and reliable execution in practice \cite{WZ-JPQ-TW:20}. 
Traditionally, bridging this gap has demanded 
extensive manual parameter tuning, ad-hoc heuristics, and exhaustive 
debugging, which are time-consuming and difficult to scale. 

Recent advancements in LLMs 
offers a promising way to synthesize controller 
directly from high-level task descriptions and interface specifications. 
Although LLMs can substantially reduce the engineering effort 
required for controller design, one-shot code generation remains unreliable.
Generated code often suffers from syntax errors, runtime crashes, 
safety violations, or a fails achieve task success. These failures often arise from 
underspecified prompts that omit important details 
such as robot dynamics, unit conversions, timing constraints, and 
API contracts. In the absence of these grounded details, LLMs frequently 
fall back on hallucinated assumptions that do not hold in specific hardware 
environments, resulting in brittle performance.

To address these challenges, we introduce a closed-loop, 
agentic workflow that combines LLM-based synthesis 
with automated verification and repair. Our framework transforms 
high-level requirements into candidate controller for a robot, which are then 
evaluated with a comprehensive test suite assessing 
the performance. Diagnostic feedback from these evaluations enables 
iterative refinement of the prompt or directly the code, 
ensuring the final controller satisfies all safety 
and performance requirements before deployment.
This paradigm shifts the engineering effort from 
manual implementation to an automated workflow.

\subsection{Related Work}\label{sec:related}

\textbf{LLM enabled robotics.}  Recent research have used LLMs as 
high-level decision-makers to enhance autonomy in domains such 
as autonomous driving \cite{RS-WZ-WF-TF-SR-WYW:23,
BL-YW-JM-BI-SV-KL-MP:24,SPS-FP-VK-MC:23}, service robotics 
\cite{KR-JH-SG-JA-IR-NS:23,DH-MB-FD-TW-AV:24,ZH-FL-CS-YS-AF-SM-AG-JB:24}, 
and system diagnosis \cite{RS-AE-CA-MF-ES-MP:24,AT-KK-TZ-MP-CTT-JPH:24}. 
In these works, LLMs typically function as task planners that decompose 
instructions into logical subgoals, and use pre-existing low-level controllers 
for execution. 
However, using LLMs for direct low-level control remains a significant challenge, 
as these models often lack the precision and safety guarantees required for
tasks such as trajectory generation and path planning \cite{MA-EP-ZY:24}. While techniques such as prompt 
enrichment \cite{CHS-JW-CW-BMS-WLC-YS:23, ML-SA-MSK:24}, Retrieval-Augmented 
Generation (RAG) \cite{PL-EP-AP-FP-VK-NG-HK-ML-WTY-TR:20}, and environment-aware 
refinement \cite{MJ-SG-SM-YM-ORZ:25} have improved reasoning consistency, they often 
use an LLM during execution. This dependency introduces latency and potential 
stochastic failures at runtime. In contrast, our framework utilizes 
environment feedback to synthesize a standalone, executable controller. Once 
verified through our agentic loop, the resulting controller is deployable without further 
need of an LLM to correct the robot controller. 

\textbf{LLM-based coding agents.} A growing body of work explores LLMs as 
coding agents for robotic control. These frameworks typically follow a loop:
\texttt{code $\rightarrow$ execute $\rightarrow$ verify $\rightarrow$ replan $\rightarrow$ code}.
For example, \emph{Code as Policies} \cite{JL-WH-FX-PX-KH-BI-PF-AZ:22} enables modular program synthesis through recursive function generation. Other methods incorporate runtime assertions \cite{IS-VB-AM-AG-DX-JT-DF-JT-AG:22} or classical optimization-based controllers, such as MPC \cite{GM:25}, to inform subsequent code regeneration. In contrast, we introduce an offline, test-driven code generation framework centered on structured PyTest suites. Using failure diagnostics to iteratively refine prompts or perform direct source-level edits prior to deployment, our approach produces a self-contained controller that operates without external solvers or specialized libraries.

\textbf{Agentic workflows.} Recent work on LLM-based robotics is 
moving from open-loop reasoning toward closed-loop, self-correcting 
execution \cite{AB-YC-CF-KH-AH-DH-JI-AI-EJ-RJ:23}. A key 
mechanism in this shift is verbal learning 
\cite{NS-FC-AG-KN-SY:23, SS-FP-MC:23}, where the agent reflects 
on failure cases in natural language and uses the resulting critique to 
improve subsequent decisions. Our approach is closely related in spirit 
to verbalized machine learning \cite{TZX-RB-BS-WL:25}, which interprets 
language feedback as an update signal analogous to a gradient step. 
Building on this idea, we propose a dual-LLM design: a Learner 
that generates the robot controller, and an Optimizer that 
performs iterative improvement by refining prompts or directly editing code.

\subsection{Contributions}
The main contributions of this paper are as follows:
\begin{itemize}
    \item We propose a novel framework that leverages an agentic workflow to synthesize robotic controllers from under-specified high-level instructions. In contrast to one-shot prompting, our approach incorporates an automated, test-driven refinement loop that iteratively improves the generated controller, ensuring syntactic correctness and compliance with task requirements.
    
    \item We present a unified methodology for controller synthesis in two representative settings:
    \begin{enumerate}
        \item \textbf{2D map-based navigation:} A pipeline that converts raw map images into occupancy grids, which are then used for controller synthesis.
        \item \textbf{3D simulation environments:} A simulator-interfacing pipeline in which the environmental metadata is used to synthesize controller.
    \end{enumerate}
    
    \item We validate the proposed framework through experiments in both 2D map-based navigation and 3D Webots simulation environments. 
\end{itemize}

 \section{Problem Formulation}\label{sec:formulation}
We study a robot navigation task in an obstacle-populated environment, where 
the objective is to synthesize an executable controller program, \texttt{controller.py}, that 
drives the robot from a given start position to a designated goal state while ensuring collision 
avoidance. 

Unlike classical controller design settings, the controller logic is not specified a priori. Instead, it must be generated from high-level natural-language instructions through an LLM-based agentic workflow. This introduces a key challenge: LLM-generated code is inherently stochastic and highly prompt-sensitive. Even under identical task conditions, repeated generations may produce controllers with widely varying quality. These outputs can fail at multiple levels, including, code failures (e.g., syntax errors, 
import errors, runtime exceptions) and performance failures (e.g., collisions, deadlock/stagnation, failure to reach the goal). Such variability makes one-shot code generation unsuitable for safety-critical robotic applications.

To address this, we aim to propose an agentic workflow to autonomously adjusts
the prompt or controller code through a closed-loop feedback
mechanism until the final controller satisfies all performance
and safety requirements validated by user defined test.
The overview of our framework is illustrated in Fig.~\ref{fig:OA2}.

\begin{figure}[!t]
  \centering
  \includegraphics[width=1\columnwidth,trim={0cm 0cm 0cm
    0cm},clip]{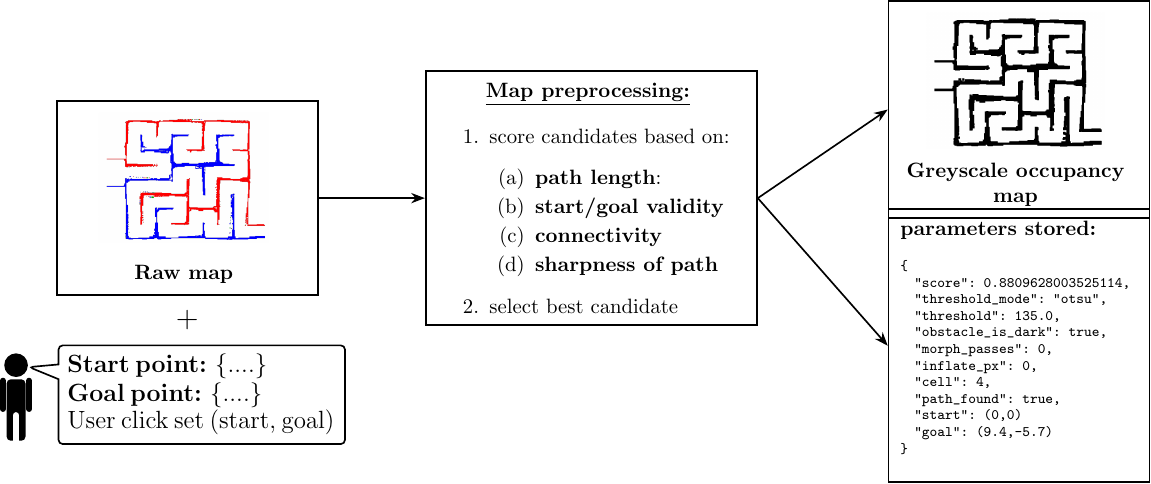}
  \caption{
The figure shows map preprocessing pipeline. A user-provided map image is 
converted into a grayscale occupancy grid $occ[y,x]$, where $occ[y,x]=\texttt{True}$ 
denotes an obstacle pixel and $occ[y,x]=\texttt{False}$ denotes free space. This 
pipeline simultaneously generates \texttt{params.json}, which stores the configuration 
parameters to produce occupancy grid. 
}
    \label{fig: OA1}
 \end{figure}


\section{Designing a Controller Agentic Workflow}\label{sec:controller_agent}
This section presents an agentic workflow for automatically 
synthesizing an executable robot controller, \texttt{controller.py} 
using task specification and environment context. The task 
specification includes the initial state, goal state, and success criteria, 
while the environment context provides the map or scene representation 
together with the robot's sensing and actuation interfaces. Given these 
inputs, the workflow autonomously produces a fully deployable 
controller without requiring any human intervention.

At a high level, the workflow follows a two-stage generate and 
improve paradigm. In the first stage, an LLM is prompted with 
the the user prompt, and environment context to generate a 
candidate \texttt{controller.py}. In the second stage, the generated 
controller is automatically validated using a \texttt{PyTest} suite that 
checks both functional correctness and safety requirements. If the 
controller fails any test, the diagnostic feedback is fed back into the 
workflow to refine the prompt and/or the generated code, and a 
new controller candidate is produced. This refinement loop continues 
until the controller satisfies all specified requirements.

We describe our methodology for two distinct settings:
a $2$D map-based environment, and a $3$D 
simulator-based environment. Although these settings differ substantially in their sensing modalities and execution backends, the underlying synthesis logic remains unchanged. In both cases, the controller is produced through the same agentic cycle of generation, evaluation, and iterative refinement.

\subsection{2D map-based environment}\label{2D}
{\bf Map preprocessing:} In the 2D setting, the user provides a raw RGB map 
image and a user prompt specifying the start and goal pixel indices. 
We transform this image into a discrete occupancy grid, $occ[y,x]\in
\{\texttt{True},\texttt{False}\}$, where \texttt{True} denotes an obstacle and \texttt{False}
denotes navigable free space. To ensure the framework remains robust 
across diverse map styles, we do not assume a priori whether obstacles 
are represented by darker or lighter pixels.

The map preprocessing pipeline follows a multi-stage transformation. 
Initially, grayscale conversion is performed by mapping the RGB input to 
a luminance-based representation \cite{ITU:15} according to the equation:
\begin{align*}
Y = 0.2126R + 0.7152G + 0.0722B.
\end{align*}
Subsequently, binarization and polarity evaluation are conducted using Otsu’s 
method \cite{NO:79}. Since the meaning of pixel intensity is unknown, we 
evaluate both binary polarities (treating dark pixels as obstacles vs. treating 
light pixels as obstacles). For each candidate polarity, we apply morphological 
operations to remove small noise, fill internal holes, and bridge narrow 
discontinuities. Obstacles are then inflated by a fixed pixel radius to provide 
a safety buffer for the robot's physical dimensions. Finally, the framework 
performs hyperparameter optimization and feasibility checks through a grid 
search over polarity, cleanup strength, and inflation radii. A candidate 
occupancy grid is deemed feasible only if the start and goal positions 
reside in free space $occ= \text{False}$ and a valid connected path exist 
between them. 

Among all feasible candidates, we select the occupancy grid that 
maximizes a weighted score favoring path efficiency (shortest Euclidean 
distance), safety (maximum clearance from obstacles), and boundary 
smoothness. and then pass the chosen grid and associated parameters 
to the controller-generation stage. This optimized grid and its associated 
metadata are then used to generate controller. The 
map-preprocessing is shown in Fig.~\ref{fig: OA1} and described 
in Algo.~\ref{alg:compact_occ_path}.

\begin{algorithm}[t]
\caption{Pseudo code for map preprocessing}
\label{alg:compact_occ_path}
\KwIn{user-given map $I$, start point $A$, goal point $B$}
\KwProc{
$G\gets \textsf{Grayscale}(I)$, $T\gets \textsf{Otsu}(G)$\;
$\mathcal{B}\gets \{dark,light\}$ \tcp*{binarization}
$\mathcal{I}\gets \{0,1,2,3\}$ \tcp*{inflate parameter}
$\mathcal{C}\gets \{0,4,8,12,16\}$ \tcp*{cleanup parameter}
$\Theta \gets {T}\times\mathcal{B}\times\mathcal{I}\times\mathcal{C}$ \tcp*{candidates}
\ForEach{$\theta\in\Theta$}{
  score$(\theta)$ = $2.5~*$ min clearance $ - 0.02~*~$path~length~$-~1.2~*~$sharpness\;
  \If{$A$ or $B$ lie on obstacles}{ score$(\theta)=-\infty$ }
  \If{$A$ or $B$ has no connectivity}{ score$(\theta)=-\infty$ }
 \KwRet{$\textsf{best}~\theta$}\;
  }
}
\KwOut{Greyscale occupancy map, parameters $\theta$}
\end{algorithm}

\begin{figure*}[!t]
  \centering
  \includegraphics[width=1\textwidth,trim={0cm 0cm 0cm 0cm},clip]{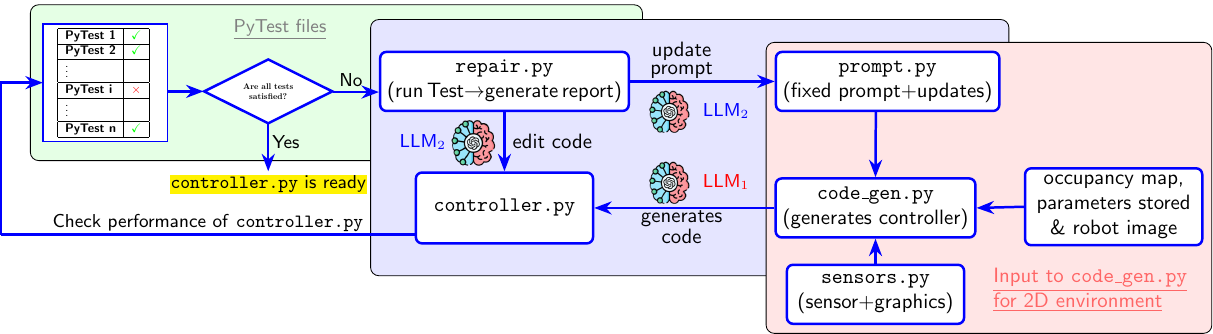}
\caption{The figure shows the overview of our test-driven agentic 
framework. The workflow is partitioned in to three functional 
domains: (1) The red block shows the inputs to \texttt{code\_gen.py}, 
providing preprocessed occupancy maps with \texttt{params.json} 
and the robot image; (2)  The blue block highlights the calibration 
loop, where \texttt{code\_gen.py} generates \texttt{controller.py} 
from \texttt{prompt.py}. \texttt{PyTest} suite evaluates the generated 
controller, and \texttt{repair.py} updates the prompt template or 
updates the code based on the failures; (3) The green block defines 
the PyTest files, which are run via \texttt{repair.py}. }
  \label{fig:OA2}
\end{figure*}

{\bf Robotic controller generation:}
Following map preprocessing, we describe an agentic workflow to synthesize an 
executable robot controller as a single Python script, \texttt{controller.py}.
An overview of this architecture is provided in Fig.~\ref{fig:OA2}. 

The synthesis module, \texttt{code\_gen.py}, takes the following inputs:
\begin{enumerate}
\item \emph{\texttt{prompt.py}} (fixed prompt and dynamic updates): 
This file provides the instructions to the LLM. Its fixed component 
(\texttt{PROMPT\_FIXED}) specifies non-negotiable requirements, such 
as the set of permitted libraries and mandatory functions that must be 
included in the generated controller. The dynamic updates consists of 
\texttt{AUTO\_REPAIR\_RULES}, which are updated at each iteration 
using diagnostics from PyTest failures. This design enables iterative 
correction and refinement without modifying the base prompt. The 
template of \texttt{prompt.py} is described in Listing~\ref{controller template}.

    \begin{figure}[t]
\centering
\begin{minipage}{0.98\linewidth}
\begin{lstlisting}[
  language=Python,
  caption={Template of \texttt{prompt.py}},
  label={controller template}
]
from string import Template

PROMPT_FIXED = Template(r"""
# write the fixed instructions
""")
AUTO_REPAIR_RULES = r"""
# === AUTO_REPAIR_RULES_BEGIN ===
# LLM writes instructions based on the PyTest results (from repair.py)
# === AUTO_REPAIR_RULES_END ===
"""
PROMPT = Template(PROMPT_FIXED.template + "\n" + AUTO_REPAIR_RULES + "\n")
\end{lstlisting}
\end{minipage}
\end{figure}
    
    \item \emph{\texttt{sensors.py} (Support module)}: A lightweight auxiliary module 
    used by the controller. It implements a \texttt{Graphics} class for Pygame-based 
    visualization and an \texttt{Ultrasonic} class for occupancy-grid ray-casting, which 
    is used to simulate range sensing.
    \item \emph{Environment context and robot assets:} The environment inputs include 
    a binary occupancy grid and a configuration file, \texttt{params.json}, which specifies 
    the map resolution, relevant parameters (e.g., threshold and polarity), and the robot 
    start/goal locations in pixel coordinates. A robot image is also provided for 
    visualization of the robot on the map.

\end{enumerate}

The \texttt{code\_gen.py} aggregates the environment and 
robot contexts, injecting them into the \texttt{prompt.py}  
to generate the candidate \texttt{controller.py}. We validate 
each candidate controller using 
a PyTest suite. 
The \texttt{repair.py} script executes the PyTest suite and generates 
a structured report that summarizes each failing test case along with 
its associated error cause. We treat this report as structured feedback and 
pass it to another LLM to drive one of two repair actions: 
\begin{enumerate}
\item \emph{Code-level repair:} directly modify the current \texttt{controller.py} 
to resolve reported failures. This local refinement persists for a set 
number of attempts until the edit patience limit is reached.
\item \emph{Prompt-level repair:} if repeated code edits do not resolve the failures
within the edit limit, apply targeted updates 
to \texttt{prompt.py}. The revised prompt is then passed to \texttt{code\_gen.py} to get 
an improved \texttt{controller.py}.
\end{enumerate}
This evaluate, repair, and regenerate loop repeats until the controller passes all tests 
or the maximum iteration threshold is reached, ensuring that the final output is 
verified and deployable. The pseudo code for this controller generation is 
provided in Algo.~\ref{alg:controller_agent}.

\begin{algorithm}[t]
\caption{Test-driven Controller Synthesis}
\label{alg:controller_agent}
\DontPrintSemicolon

\SetKwInOut{Input}{Input}
\SetKwInOut{Output}{Output}

\Input{Prompt template $\pi$, Verification suite $\mathcal{V}$, Max iterations $K$, Repair patience $J$, Environmental context $\mathcal{E}$}
\BlankLine
\KwProc{
Initialize prompt $\mathcal{P} \leftarrow \text{PutContext}(\pi, \mathcal{E})$\;

\For{$k \leftarrow 1$ \KwTo $K$}{
    $\mathcal{C} \leftarrow \text{LLM-Generate}(\mathcal{P})$ \tcp*{\small Synthesize code}
    
    \For{$j \leftarrow 0$ \KwTo $J$}{
        $\mathcal{R} \leftarrow \text{RunTests}(\mathcal{C}, \mathcal{V})$ \tcp*{\small PyTest report}
        
        \If{$\mathcal{R}.\text{passed} = \text{True}$}{
            \Return $\mathcal{C}$ \tcp*{\small Exit and deploy}
        }
        
        \BlankLine
        \uIf{$j < J$}{
            $\mathcal{C} \leftarrow \text{Edit}(\mathcal{C}, \mathcal{R})$ \tcp*{\small Try code repair}
        }
        \Else{
            $\sigma \leftarrow \text{Summarize}(\mathcal{R})$\;
            $\pi \leftarrow \text{UpdateTemplate}(\pi, \sigma)$\; 
            $\mathcal{P} \leftarrow \text{InjectContext}(\pi, \mathcal{E})$\;
        }
    }
}
}
\Output{Verified controller $\mathcal{C}$ (\texttt{controller.py})}
\end{algorithm}

\subsection{$3$D simulator-based environment}\label{3D}
We further evaluate the proposed agentic workflow in a $3$D, physics-based simulation environment using a differential-drive mobile robot equipped with infrared range sensors. Unlike the 2D map-based setting, this scenario requires no map preprocessing; instead, the synthesized controller operates directly on a simulator world configuration.

\begin{figure}[!t]
  \centering
  \includegraphics[width=0.49\textwidth,trim={0cm 0cm 0cm 0cm},clip]{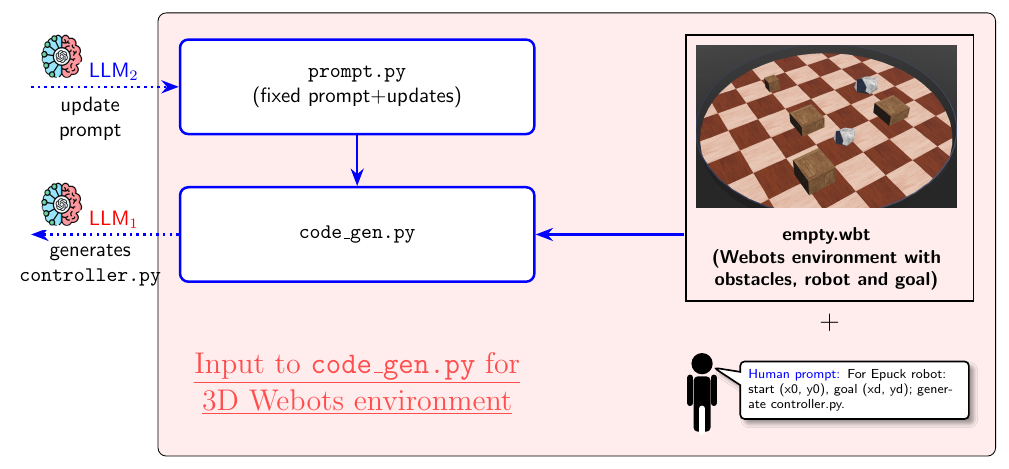}
\caption{The figure shows the inputs to \texttt{code\_gen.py} in the 
$3$D Webots simulation. The generator (LLM$_1$) produces 
\texttt{controller.py} conditioned on (i) the current prompt 
specification \texttt{prompt.py} (fixed base prompt plus 
accumulated updates), (ii) the Webots world file \texttt{empty.wbt} 
describing the arena, robot, goal, and static obstacles, and (iii) a 
human task prompt specifying start and goal poses. The optimizer 
(LLM$_2$) uses \texttt{Pytest} feedback to update \texttt{prompt.py} 
across iterations.}
  \label{fig:webots3}
\end{figure}

The agentic loop matches Fig.~\ref{fig:OA2}, except that \texttt{code\_gen.py} uses the simulator API to generate a controller for execution in Webots. Experiments use a fixed world specified by \texttt{envt.wbt}, which defines a bounded arena with a diverse set of static obstacles. The world file is provided to the LLM for grounding, including: (i) \emph{robot embodiment} (physical dimensions and differential-drive sensing/actuation interface), (ii) \emph{environment geometry} (obstacle placement and geometry), and (iii) \emph{task coordinates} (initial and goal positions in the world frame). Conditioned on this specification, the LLM produces a \texttt{controller.py} file that is executed in Webots. Figure~\ref{fig:webots3} summarizes the inputs to \texttt{code\_gen.py} in this setting.

\subsection{Performance metrics}
For each experimental setting, we perform $R$ independent runs, indexed by $r \in \{1,\ldots,R\}$. Each run proceeds for at most $K$ iterations, indexed by $k \in \{1,\ldots,K\}$, and terminates early if the generated controller passes the complete \texttt{PyTest} suite. Let $f_{r,k}$ denote the number of failing tests at iteration $k$ of run $r$, and define the success indicator $s_{r,k}=\mathbb{I}\{f_{r,k}=0\}$. 
The first-success iteration (stopping time) for run $r$ is defined as:
$\tau_r=\min\{k\in\{1,\ldots,K\}: s_{r,k}=1\}$,
with $\tau_r=\infty$ if no success occurs within $K$ iterations.
We evaluate the framework using two performance metrics:
\begin{enumerate}
\item {{Success rate (SR):}} The fraction of generated controller candidates that 
pass the complete test suite: 
\[
\mathrm{SR}=\frac{1}{|\mathcal{V}|}\sum_{(r,k)\in\mathcal{V}}\mathbb{I}\{f_{r,k}=0\},
\]
where $\mathcal{V}$ denotes the set of valid run-iteration pairs $(r,k)$ for
which a controller was generated

\item {{Cumulative success (CS):}} The fraction of runs that have succeeded 
at least once by iteration $k$, effectively representing the empirical Cumulative 
Distribution Function (CDF) of $\tau_r$:
\[
\mathrm{CS}(k)=\frac{1}{R}\sum_{r=1}^{R}\mathbb{I}\{\tau_r\le k\}.
\]
This metric summarizes progress over iterations and indicates the proportion of runs solved by the $k$-th iteration.
\end{enumerate}

\section{Experiments}\label{sec:experiments}
We evaluate the proposed framework on ground-vehicle navigation 
tasks in environments with static obstacles. Our evaluation spans
 two domains: (i) a $2$D navigation environment and (ii) a $3$D
physics-based simulation in Webots simulator.

\subsection{$2$D navigation setup}\label{subsec:2d_setup}
We first consider a planar navigation task where the robot must navigate 
a raw map image. The objective is to autonomously synthesize a 
differential-drive controller (\texttt{controller.py}) that directs the 
agent from a initial position to a goal position. 

The robot is initialized with heading $\theta_0=40$ rad. Actuator 
constraints are imposed with wheel speeds $v_l^{\max}=
v_r^{\max}=0.02$\,px/s. To ensure safety, the robot is 
equipped with local ultrasonic sensors for obstacle detection. 

\paragraph{Environment and guidance signal}
As detailed in Sec.~\ref{sec:controller_agent}, the raw map 
image is pre-processed into a grid-occupancy representation
using Algo.~\ref{alg:compact_occ_path}. To ensure task feasibility 
an $A^*$ search algorithm computes a reference route from the 
start to the goal. This serves as connectivity check for controller synthesis.
For real-time monitoring and debugging of the robot's kinematics, we utilize 
a custom \texttt{PyGame}-based visualizer to visualize the robot's motion.

\paragraph{Task execution and iterative repair}
The controller generation follows the agentic loop illustrated in 
Fig.~\ref{fig:OA2}. The fixed prompt is
illustrated in Listing~\ref{lst:controller_template1}, and the 
the prompt is dynamically updated based on 
\texttt{PyTest} failures (see Appendix for examples). 
The repair process has code-level repair and prompt-level repair. We 
consider the optimizer LLM to have an edit patience of $J=1$ 
attempts. If the controller fails to 
satisfy the verification suite after $J$ attempts, the system updates 
the \texttt{AUTO\_REPAIR\_RULES} to refine the prompt for 
subsequent iterations. 

\paragraph{Observations}
We perform $R=5$ independent runs per model, with a maximum budget of $K=20$
iterations per run. With the optimizer LLM is fixed (\mbox{GPT-4o}), 
we evaluate the performance of four different learner LLMs:
\mbox{GPT-4o}, \mbox{GPT-4.1}, \mbox{GPT-5.2}, \mbox{Claude-Sonnet-4.5} 
and \mbox{Claude-Opus-4.5}. Success is measured via the 
\texttt{PyTest} suite in Table~\ref{tab:pytest}.

\begin{figure}[!t]
  \centering
  \includegraphics[width=0.48\textwidth,trim={0cm 0cm 0cm 0cm},clip]{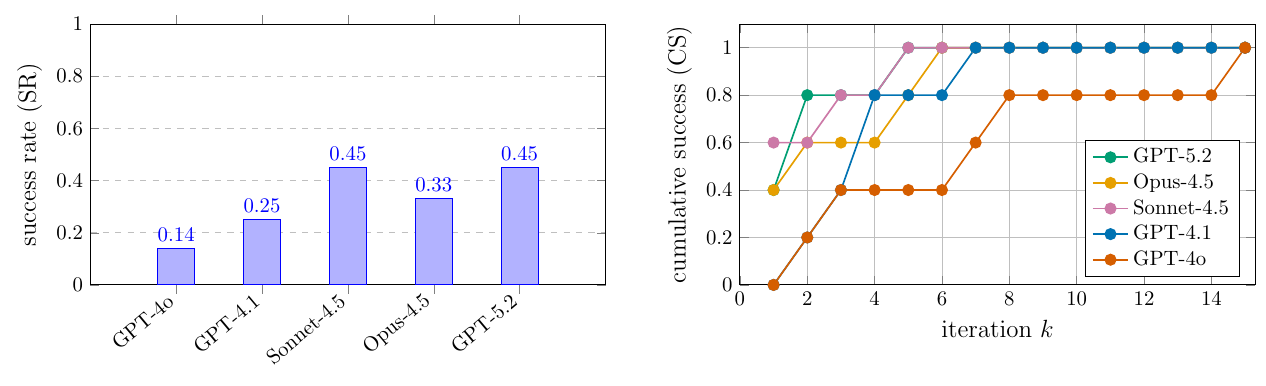}
\caption{The figure shows the performance of our approach in $2$D navigation described in Section~\ref{subsec:2d_setup}. The left 
figure shows the success rate across learner models. The right figure shows cumulative 
success vs iteration $k$. 
}
  \label{fig:success_rate}
\end{figure}

The performance of different models are summarized in Fig.~\ref{fig:success_rate}. 
We observe that \mbox{GPT-5.2} and \mbox{Sonnet-4.5} has a success rate (SR) 
of $45\%$, followed by \mbox{Opus-4.5} $(33\%)$,  \mbox{GPT-4.1} $(25\%)$ 
and \mbox{GPT-4o} $(14\%)$. 
When comparing our iterative framework against a baseline 
single-shot prompting approach, our method ensures that a
deployable controller is generated by iterative refinement as in Algo.~\ref{alg:controller_agent}. 
Specifically, \mbox{GPT-4o} and \mbox{GPT-4.1} failed 
to reach success in a single iteration across all experimental runs, 
\mbox{Opus-4.5} and \mbox{GPT-5.2} succeed in 
$2$ iterations and \mbox{Sonnet-4.5} succeeds in $3$ iterations 
out of $5$ iterations on the first attempt. While the proposed iterative 
refinement strategy, \mbox{GPT-5.2} and \mbox{Sonnet-4.5} 
reaches $\text{CS}=1$ by iteration $k=5$, 
whereas \mbox{Opus-4.5}, \mbox{GPT-4.1}, and \mbox{GPT-4o} 
require $k=6$, $k=7$, and $k=15$ iterations, respectively.

\begin{figure}[!t]
  \centering
  \includegraphics[width=0.48\textwidth,trim={0cm 0cm 0cm 0cm},clip]{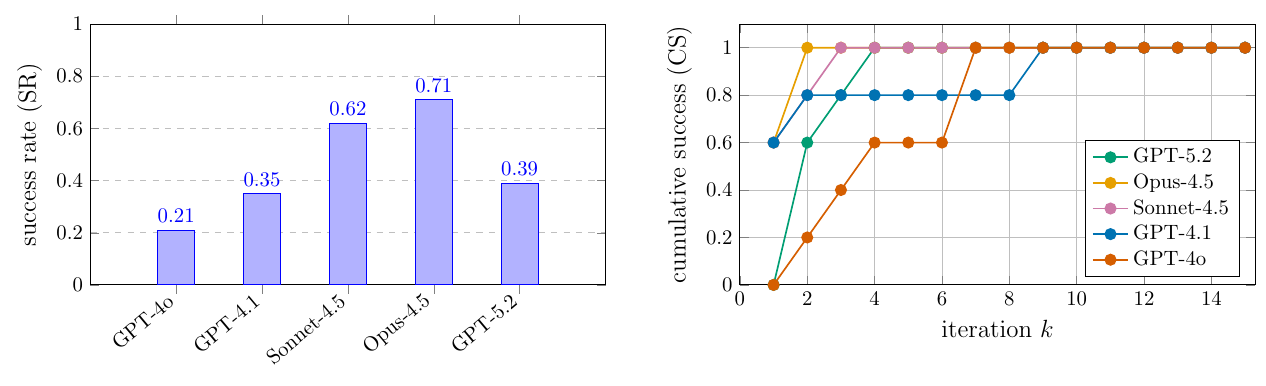}
\caption{The figure shows the controllers performance for a $3$D navigation task in a 
Webots environment described in Section~\ref{3D-Webots}. The left figure shows the success rate across learner models. 
The right figure shows cumulative success vs iteration $k$.  }
  \label{fig:success_webots}
\end{figure}

\begin{figure}[t]
\centering
\begin{minipage}{0.98\linewidth}
\begin{lstlisting}[
  language=Python,
  caption={Fixed prompt in \texttt{prompt.py} for $2$D navigation setup},
  label={lst:controller_template1}
]
PROMPT_FIXED = Template(r"""
Context (verbatim): sensors.py -> $ROBOT_PY_TEXT; params.json -> $PARAMS_JSON_TEXT
Template vars: MAP_IMG, PARAMS_JSON, ROBOT_IMG, AXLE_LENGTH_PX, SENSOR_RANGE_PX, N_RAYS
Inputs: occupancy.png (map), DDR.png (robot)

Goal / tests:
- Imports at top-level only (use allowed imports)
- is_occupied -> Python bool
- nearest_free -> Python (int,int)
- A* path -> free-space pixel waypoints
- Progress in 400 steps: dmin <= 0.7*d0
- Reach goal in 2500 steps: <20 px, collisions = 0

Setup:
- Initialize graphics + map
- Build boolean occupancy grid from map image
- Load params with safe defaults; clamp ranges
- If occupied ratio is implausible, flip polarity

Start/goal selection:
- Deterministic candidates (fixed RNG + corners)
- Snap start/goal to nearest free

Control (episode loop):
- Forward speed +ve, reduce for large error
- Obstacle handling via occupancy check
- If blocked: stop motion; turn left or right
- If stuck: brief forced turn, then resume pursuit

Logging / termination:
- Record position + speed each step
- Keep collisions at zero
- Stop when within 20 px of goal
""")
\end{lstlisting}
\end{minipage}
\end{figure}

\begin{figure}[t]
\centering
\begin{minipage}{0.98\linewidth}
\begin{lstlisting}[
  language=Python,
  caption={Fixed prompt in \texttt{prompt.py} for $3$D Webots navigation setup},
  label={lst:controller_template2},
  breaklines=true,
  mathescape=false
]
PROMPT_FIXED = Template(r"""
Generate one Python Webots controller for E-puck  (code only).
- Use Supervisor. read envt.wbt. Move in XY plane.
- Goal -> Solid named "GOAL". Obstacles -> WoodenBox(size), Rock(scale approx).
- Build occupancy grid; plan 8-connected A*; follow waypoints using true translation+rotation (no dead-reckoning).
- Stop within goal tolerance (within 20 mm).
- Use main; call under __main__.
- Imports: module level only re/heapq/math; import Supervisor only inside run_episode/main.

World text: $WORLD_WBT_TEXT
""")
\end{lstlisting}

\end{minipage}
\end{figure}

\subsection{$3$D Webots simulation}\label{3D-Webots}
Next, we evaluate the proposed framework in a $3$D \textit{Webots} simulator 
environment developed by Cyberbotics. Webots is a physics-based robotics 
simulator that provides a realistic $3$D environment for developing and 
testing control, perception, and navigation algorithms before deploying 
them on hardware. It includes a library of common robot models (including 
differential-drive platforms such as the e-puck), set of sensors (e.g., proximity, 
IMU, GPS, cameras, LiDAR) and actuator models. 

\paragraph{Environment and sensing signals}
We consider an e-puck differential-drive robot equipped with eight infrared (IR) proximity 
sensors \texttt{ps0--ps7}. The controller uses six forward facing sensors, \texttt{ps0--ps2} 
for left and \texttt{ps5--ps7} for right side detection. The control is executed 
in discrete time with a timestep of $32$ ms. At each step, the controller reads 
the sensor measurements and outputs angular velocity commands for the left and 
right wheel motors, subject to a saturation limit of $6.28$ rad/s. The simulator 
provides a $3$D visualization of the scene.

\paragraph{Task execution and iterative repair}
We execute the synthesis loop using the agentic workflow 
described in Sec.~\ref{3D}, with fixed prompt as 
Listing~\ref{lst:controller_template2}. As in the 2D setting, 
we use edit patience of $J=1$. An iteration is considered 
successful when the system gives a deployable controller
before the fixed time horizon. The \texttt{PyTest} suite used 
to evaluate the controller shown in Table~\ref{tab:pytest1}.

\paragraph{Observations}
We consider $R=5$ independent runs per model, with $K=20$
iterations per run. Then, we evaluate the performance of different
LLMs as in the previous example. 
The performance of different models on the Webots simulator are summarized in 
Fig.~\ref{fig:success_webots}. We observe that 
\mbox{Claude-Opus-5.2} achieves a SR of $71\%$, 
significantly outperforming \mbox{GPT-5.2} $(39\%)$, 
\mbox{Claude-Opus-5.2} $(62\%)$,  \mbox{GPT-4.1} $(35\%)$ 
and \mbox{GPT-4o} $(21\%)$. 
We also observe that the single shot controller generation has 
a success rate of $60\%$ in \mbox{Opus-4.5}, \mbox{Sonnet-4.5}, 
and \mbox{GPT-4.1}, while other models do not succeed in first attempt.
While using the proposed iterative approach, \mbox{Opus-4.5} 
reaches CS$=1$ in $k=2$ iterations, followed by \mbox{Sonnet-4.5} $(k=3)$, 
\mbox{GPT-5.2} $(k=4)$, \mbox{GPT-4.1} $(k=9)$ and 
\mbox{GPT-4o} $(k=7)$ to converge to a reliable controller. 
We further observe that performance in the 3D Webots environment is 
better than in the 2D navigation setup, since the provided environment 
reduces the amount of computation and the unknown parameters.

\begin{table*}[h!]
\centering
\caption{PyTests for $2$D map-based simulation}
\label{tab:pytest}
\begin{tabular}{p{0.10\linewidth} p{0.10\linewidth} p{0.72\linewidth}}
\toprule
\textbf{Category} & \textbf{Test Target} & \textbf{Verification Objective} \\ \midrule
\multirow{2}{*}{\textbf{Static Contract}} & Compliance & Ensures required constants (\texttt{SENSOR\_RANGE}, etc.), imports, and \texttt{main} entry points are present. \\
 & Hygiene & Explicitly bans forbidden libraries (e.g., \texttt{scipy}) and unstable code patterns (e.g., indexing \texttt{pygame} surfaces). \\ \midrule
\multirow{3}{*}{\textbf{Unit API}} & OccMap2D & Validates boundary handling, out-of-bounds indexing, and occupancy querying logic. \\
 & Utility & Confirms \texttt{nearest\_free} correctly identifies navigable points for robot initialization. \\
 & Planner & Verifies path format (Python \texttt{int}), connectivity, and safety (no points in obstacles). \\ \midrule
\multirow{3}{*}{\textbf{End-to-End}} & Schema & Validates return signatures of simulation loops for data logging and metrics calculation. \\
 & Stability & Executes short rollouts to detect runtime exceptions or logical crashes under nominal conditions. \\
 & Success & Evaluates task completion through collision avoidance and goal-reaching within positional tolerances. \\ \bottomrule
\end{tabular}
\end{table*}

\begin{table*}[ht]
\centering
\caption{PyTests for Webots simulation}
\label{tab:pytest1}
\begin{tabular}{p{0.10\linewidth} p{0.10\linewidth} p{0.72\linewidth}}
\toprule
\textbf{Category} & \textbf{Test Target} & \textbf{Verification Objective} \\ \midrule
\multirow{2}{*}{\textbf{
\!\!Static Contract}} & {Compliance} & Ensures required constants (\texttt{SENSOR\_RANGE}, etc.), imports, and \texttt{main} entry points are present. \\
 & {Hygiene} &  Explicitly bans forbidden libraries (e.g., \texttt{scipy}) and unstable code patterns (e.g., indexing \texttt{pygame} surfaces). \\ \midrule
\multirow{3}{*}{\parbox{2cm}{\textbf{Component Logic}}} & {Environment} & Validates environment parsing from world files. \\
 & {Planner} & Verifies $A^*$ path connectivity, and ensures all planned waypoints lie strictly within free-space coordinates. \\
 & {Control} & Check accuracy of differential-drive wheel speed commands for forward motion, turning, and waypoint tracking. \\ \midrule
\multirow{2}{*}{\parbox{2cm}{\textbf{End-to-End}}} & {Schema} & Validates simulation by checking return schemas for positions, speeds, and collision counts. \\
 & {Success} & Checks performance based on iterative goal progress (30\% distance reduction), no collisions and reaches goal. \\
\bottomrule
\end{tabular}
\end{table*}

\section{Conclusion}
We proposed a test-driven agentic framework that transforms high-level 
instructions into deployable robot controller for a robot navigation task. 
By using an automated verification loop using PyTests, our approach 
gives a reliable controller which can directly be deployed on the robot. 
Our approach iteratively refines the controller code and prompt context 
until task requirements are met. Experimental validation within $2$D and 
$3$D Webots environments confirms that this closed-loop methodology 
consistently yields functional controllers within a few iterations. Ultimately, 
this framework shifts the robotics engineering paradigm from manual code 
implementation toward automation, providing a robust robot controller. 
Future research will explore the extension of this work to a dynamic 
environment with moving obstacles. 

\bibliographystyle{unsrt}
\bibliography{alias,Main,FP,New}

\begin{appendix}\label{appendix}
\section{Iterative refinement of prompt-embedded auto-repair rules}

Our prompt is refined iteratively by incorporating auto-repair rules derived from observed failures. Due to space constraints, we report only the $2$D simulation results for a single GPT-5.2 run, shown in Fig.~\ref{fig:controller-iter}; the same procedure extends to other models and to the $3$D Webots setting.

From Fig.~\ref{fig:controller-iter}, we observe that across iterations the framework repeatedly finds common issues such as API mismatches, unstable motion, and safety-contract violations. Each failure is then converted into a clear repair rule and added in the prompt for the next iteration. These rules persist across iterations and help prevent the same mistakes.
In the first iteration, our framework identified a \texttt{TypeError} caused by a deprecated argument name and automatically introduced a rule to standardize function signatures (e.g., replacing \texttt{max\_r} with \texttt{max\_rad}). In the second iteration, the controller made limited progress because it became trapped in locally stable configurations. The framework responded by adding an ``unstick'' maneuver to the prompt, improving robustness in simulation. Overall, the prompt evolves to match the environment and past failures, so the agent learns better strategies over time.

\begin{figure*}[t]
\centering
\begin{minipage}{\textwidth}
\centering

\begin{StageBox}
    \vspace{-1mm}
\noindent\textbf{\underline{Iteration $k=1$:} }\\
\begin{center}
\vspace{-4mm}
\textit{Generate \texttt{controller.py} using fixed prompt \\$\Downarrow$\\}
\vspace{-2mm}
\end{center}

\begin{TaskBox}{Diagnostic Report (Iteration $k=1, e=0$)}
    \vspace{-2mm}
\begin{enumerate}
    \item The robot did not reach the target ($d_{\min}=102.5$, collisions$=0$).
    \item Found forbidden \texttt{scipy} library dependency. Test setup aborted.
    \item Script contains prohibited \texttt{Runtime Error} raising. Test setup aborted.
\end{enumerate}
    \vspace{-4mm}
\end{TaskBox}

\begin{center}
\vspace{-3mm}
    $\Downarrow$\\ \textit{Edit \texttt{controller.py} {(no prompt update)}}\\ $\Downarrow$\\
    \vspace{-2mm}
\end{center}

\begin{TaskBox}{Diagnostic Report (Iteration $k=1, e=1$)}
    \vspace{-2mm}
\begin{enumerate}
    \item No significant progress toward goal ($209.6 (d_{min}) \geq 0.7 \times 222.4 (d_0)$).
    \item Did not reach goal. ($d_{\min}=199.9$, \texttt{collision}=0).
    \item Found forbidden \texttt{scipy} library dependency. Test setup aborted.
    \item Script contains prohibited \texttt{Runtime Error} raising. Test setup aborted.
\end{enumerate}
    \vspace{-4mm}
\end{TaskBox}
\begin{center}
\vspace{-3mm}
    $\Downarrow$\\ \textit{Edit patience reached}
    \vspace{-1mm}
\end{center}
\end{StageBox}

\vspace{2mm}

\begin{StageBox}
    \vspace{-1mm}
\noindent\textbf{\underline{Iteration $k=2$:}}\\

\begin{center}
    \vspace{-5mm}
    \textit{Update prompt}\\ $\Downarrow$
        \vspace{-2mm}
\end{center}

\begin{EnvBox}{New rules added to the fixed prompt}
    \vspace{-2mm}
\begin{itemize}[leftmargin=1.4em,itemsep=1mm,topsep=1mm]
    \item Accept both argument (\texttt{max\_rad} and legacy \texttt{max\_r}) to avoid \texttt{TypeError}.
    \item Guarantee the returned cell is free by validating occupancy and expanding the search radius up to \texttt{max\_rad}.
    \item Default \texttt{vel=3.2} for \texttt{goal\_dis>350}; allow \texttt{4.2} when clearance and heading error are small; cap to \texttt{2.4} for less clearance.
    \item If \texttt{min\_front\_range<0.45}, stop and turn; if \texttt{<0.30}, turn toward the more open side.
    \item While \texttt{avoid\_steps>0}, execute a fixed turning maneuver and suppress go-straight overrides.
    \item Track best goal distance; if no improvement for \texttt{25} steps, trigger an unstick mode for \texttt{40} steps.
    \item If clearance drops below threshold, revert to a slow turn toward open space.
\end{itemize}
    \vspace{-4mm}
\end{EnvBox}

\begin{center}
    \vspace{-3mm}
    $\Downarrow$\\ \textit{Generate \texttt{controller.py}}\\ $\Downarrow$\\
        \vspace{-2mm}
\end{center}

\begin{TaskBox}{Diagnostic Report (Iteration $k=2, e=0$)}
    \vspace{-2mm}
\begin{enumerate}
    \item Found forbidden \texttt{scipy} library dependency. Test setup aborted.
    \item Script contains prohibited \texttt{Runtime Error} raising. Test setup aborted.
\end{enumerate}
    \vspace{-4mm}
\end{TaskBox}

\begin{center}
    \vspace{-3mm}
    $\Downarrow$\\ \textit{Edit \texttt{controller.py} (no prompt update) }\\ $\Downarrow$\\
        \vspace{-1mm}
\end{center}

\begin{center}
    \vspace{-2mm}
        \textit{All PyTests passed {\color{green!60!black}\checkmark} (controller ready to deploy)}\\
    \textbf{Output: \texttt{controller.py} }
\end{center}
\end{StageBox}

\end{minipage}
\caption{The figure shows the prompt-repair iterations in the $2$D simulation for a single run using GPT-5.2. Each iteration updates the prompt with new auto-repair rules derived from the previous failure, leading to a deployment-ready controller.}
\label{fig:controller-iter}
\end{figure*}


\end{appendix}

\end{document}